\documentclass[letterpaper, 10 pt, conference]{ieeeconf}  

\usepackage{times}
\usepackage{xcolor}
\usepackage{graphicx}
\usepackage{amsmath}
\usepackage{amssymb}
\usepackage{booktabs}
\usepackage{url}
\usepackage{multirow}
\usepackage{url}
\usepackage[space]{cite}
\usepackage[section]{placeins}
\usepackage[breaklinks=true,bookmarks=false]{hyperref}
\graphicspath{ {./images/} }
\usepackage[margin=0.792in]{geometry}
\usepackage{tikz}
\usepackage{comment}
\usepackage{color}
\usepackage{gensymb}

\IEEEoverridecommandlockouts                              

\title{\LARGE \bf
SGL: Structure Guidance Learning for Camera Localization
}

\author{Xudong Zhang$^{1}$, Shuang Gao$^{1}$, Xiaohu Nan$^{2}$, Haikuan Ning$^{3}$, Yuchen Yang$^{1}$, Yishan Ping$^{1}$, \\
Jixiang Wan$^{1,4}$, Shuzhou Dong$^{1}$, Jijunnan Li$^{1}$, Yandong Guo$^{1}$
\thanks{$^{1}$ OPPO Research Institute, Shanghai, China. \{zhangxudong, gaoshuang, yangyuchen, pingyishan, wanjixiang, dongshuzhou1, lijijunnan, guoyandong\}@oppo.com}
\thanks{$^{2}$ Xiaohu Nan is with University of Chinese Academy of Sciences, Beijing, China.}
\thanks{$^{3}$ Haikuan Ning is with Wayz Intelligent Manufacturing Technology Co. Ltd., Wuxi, China.}
\thanks{$^{4}$ Jixiang Wan is also with Department of Automation, Shanghai Jiao Tong University, Shanghai, China.}
}

\begin{document}

\maketitle
\thispagestyle{empty}
\pagestyle{empty}

\begin{abstract}

Camera localization is a classical computer vision task that serves various Artificial Intelligence and Robotics applications. With the rapid developments of Deep Neural Networks (DNNs), end-to-end visual localization methods are prosperous in recent years. In this work, we focus on the scene coordinate prediction ones and propose a network architecture named as Structure Guidance Learning (SGL) which utilizes the receptive branch and the structure branch to extract both high-level and low-level features to estimate the 3D coordinates. We design a confidence strategy to refine and filter the predicted 3D observations, which enables us to estimate the camera poses by employing the Perspective-n-Point (PnP) with RANSAC. In the training part, we design the Bundle Adjustment trainer to help the network fit the scenes better. Comparisons with some state-of-the-art (SOTA) methods and sufficient ablation experiments confirm the validity of our proposed architecture.

\end{abstract}

\section{INTRODUCTION}

Camera Localization aims to estimate the accurate position and orientation with the input image, which is a crucial and fundamental computer vision task in many Artificial Intelligence and Robotics applications, such as Autonomous Driving and AR/VR. Camera Localization also works as a key component in the Visual Simultaneous Localization and Mapping (V-SLAM) and Structure From Motion (SFM) algorithms. Traditionally, structure-based camera localization methods rely on the multi-view geometry theory to calculate the 6-DoF poses by extracting and matching the feature points between the input images and the retrieved database images. Generally, Image Retrieval Convolution Networks, e.g., NetVLAD \cite{arandjelovic2016netvlad}, DELF \cite{noh2017large} and DELG \cite{cao2020unifying}, are widely adopted to obtain the high-quality global features of every image  to accomplish the retrieval task. Moreover, in the past decades, numerous local-feature descriptors including hand-craft ones, e.g., SIFT \cite{valgren2007sift}, SURF \cite{murillo2007surf} and ORB \cite{rublee2011orb}, and learning-based ones, e.g., SuperPoint \cite{detone2018superpoint}, R2D2 \cite{revaud2019r2d2} and DISK \cite{tyszkiewicz2020disk} have been proposed to get more precise and stable 2D-3D matches. Commonly, the corresponding 3D points are generated by SFM, V-SLAM, or additional depth sensors. Given the above, 6-DoF Camera poses will be recovered by PnP \cite{gao2003complete} with RANSAC \cite{fischler1981random} ultimately. The described traditional structure-based camera localization pipeline has attracted dominated research interests in recent years, while there are still some challenging corner cases remaining challenging, such as illumination changes, texture-less scenes and repetitive structures.

\begin{figure}
	\includegraphics[width=8cm]{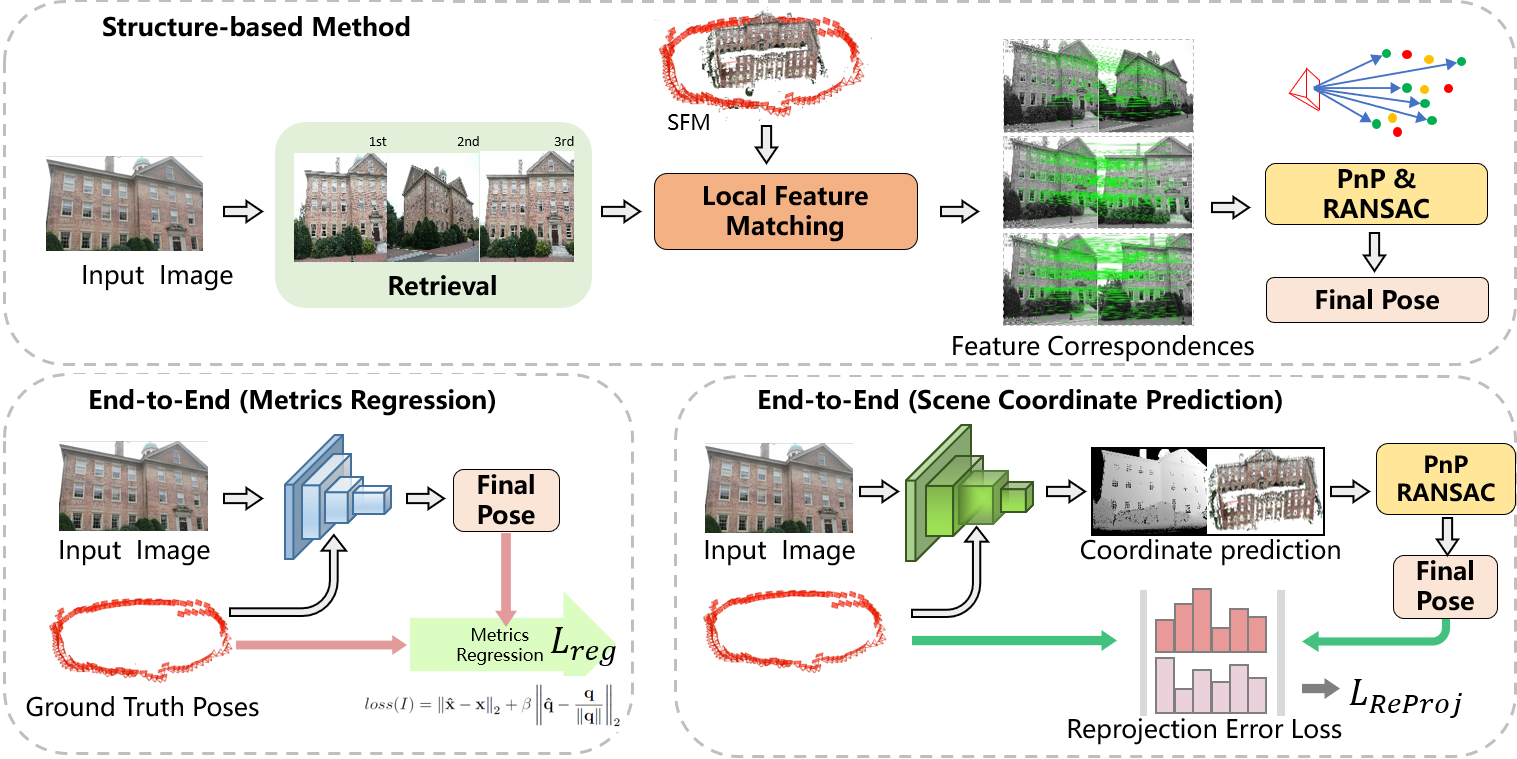}
	\centering
	\caption{
		\textbf{Demonstration of different camera localization architectures.} \textbf{Upper}: Classical Structure-based camera localization pipeline. SFM and Image Retrieval part take charge of getting image pairs and 3D point clouds. By applying the local feature matching method, feature correspondences are established, leading to 2D-3D matches. Ultimately, PnP + RANSAC will ensure us the stable estimated pose. \textbf{Lower left}: Metrics Regression style end-to-end localization pipeline. In this branch, DNNs are trained only to regress the metrics (position and orientation) without explicit relations to 3D structure information. \textbf{Lower Right}: Scene Coordinate Prediction style end-to-end localization pipeline. This branch of methods applies DNNs to predict key-points 3D position in the global frame, and use PnP + RANSAC to estimate the final pose. In the model training stage, reprojection error loss is favoured usually.
	}
	\label{fig:introduction}
\end{figure}

Due to the recent advances of Deep Neural Networks (DNNs) in the computer vision tasks, a variety of researches report the end-to-end camera localization approaches, which estimate the camera poses directly through the neural networks inference without the additional 3D model or depth information. Different from structure-based methods, end-to-end ones replace the mapping stage (e.g. SFM) with the network training process. Essentially, during the training, the models encode the scenes and represent them with the trained weights instead of 3D point clouds in structure-based methods. With the input image and intrinsic parameters, the network produces the final estimated pose through network inference. According to the training pattern, end-to-end also can be separated into two branches, metrics regression (directly regressing the position and orientation vectors) and scene coordinates prediction (predicting the corresponding 3D positions with respect to the 2D key-points and estimate the poses with PnP). Different architectures (shown in Fig. \ref{fig:introduction}) lead to distinct localization performances. The common target of most of the camera localization methods is to address the most accurate positions and orientations.

In this paper, we propose an end-to-end framework camera localization method, named as Structure Guidance Learning (SGL) Network. Basically, we take the advantage of DNNs to estimate the key-points 3D positions of the input image in the global coordinate frame, therefore producing the 2D-3D correspondences. Consequently, we are able to calculate the camera poses by PnP and RANSAC. In order to pursue the high performance, we not only carefully design and tune the network architecture, but also refer to the experience and technique in structure-based localization approaches like Bundle Adjustment (BA) \cite{konolige2008frameslam} and Key-point filtering \cite{royer2017benchmarking}, and merge them into our network training pipeline. Hence, our proposed method holds both advantages of structure-based and end-to-end approaches, and outperforms some SOTA camera localization methods on open-source datasets. Our contributions are summarised as follows:

1. We propose an end-to-end scheme camera localization method, called Structure Guidance Learning Network (SGL). With sufficient experiments, our method outperforms some SOTA ones on the prevailing open-source indoor and outdoor localization datasets, including Microsoft 7Scenes \cite{shotton2013scene} and Cambridge Landmarks Dataset \cite{kendall2015posenet}.

2. We apply the bilateral network structure to extract and combine the high-level and low-level visual features. In the confidence branch, we format the confidence map and select the high-quality key-points and conduct PnP to calculate the pose afterwards. By merging the confidence branch with the structure branch, our model functions as a point-wise attention model \cite{hu2018squeeze} which succeed in reaching the high accuracy performances.

3. Inspired by the structural idea in traditional BA, we develop a novel training technique that feeds back the reprojection error of the retrieved images and their key-point correspondences. Accordingly, we utilize the additional retrieval and image-matching model to help us to make the end-to-end network fit the scenes better.

\section{Related works}

\label{related_works}
In this section, we discuss the related works on the different camera localization frameworks, including the traditional structure-based methods, the end-to-end metrics regression ones and the end-to-end scene coordinates prediction ones.

\textbf{Structure-based Localization.} The core of structure-based method is how to obtain high-quality matching pairs between 2D features in the query image and 3D points in the SFM reconstruction model, and then get the camera poses according to the perspective geometry theory \cite{zhou2021retrieval} \cite{gao2022pose}. \cite{li2010location} compares the descriptors of the query features and 3D points directly. It works well in some small-scale scenes, but faces some drawbacks in some large ones. \cite{5206587} draws on the image retrieval techniques to narrow down the database, and obtains 2D-3D correspondences indirectly through matching 2D features between the query image and database images. It may cause losses of matching information due to different views angles. And \cite{6599055} \cite{zhang2020reference} extends the database volumes by generating the rendered synthetic images as the database. Active Search \cite{7572201} implements the image retrieval process followed by the direct 2D-3D matching, so it covers both advantages. On these bases, the applications of some excellent feature extraction \cite{detone2018superpoint} \cite{revaud2019r2d2}, feature matching \cite{sarlin2020superglue} and image retrieval \cite{noh2017large} \cite{cao2020unifying} methods can further improve the camera localization performances individually.

\textbf{End-to-End Metrics Regression.} Metrics regression localization methods aim to regress the camera pose directly from train images with the ground truth poses. PoseNet \cite{kendall2017geometric} \cite{kendall2015posenet} trains a CNN to regress the 6-DOF camera pose from a single RGB image without additional engineering or graph optimization. And it has been extended to video mode using LSTM to extract temporal information \cite{walch2017image}. Later on, \cite{kendall2016modelling} uses a Bayesian CNN implementation to obtain an estimate of the localization uncertainty and improves the accuracy on the large-scale outdoor datasets. AtLoc \cite{wang2020atloc} shows that the attention block can be used to force the network to focus on more geometrically robust objects and features, which can learn to reject dynamic objects and illumination conditions to achieve better performance. MapNet \cite{brahmbhatt2018geometry} exploits other sensory inputs like visual odometry and GPS in addition to images, and fuses them together for camera localization.

\textbf{End-to-End Scene Coordinates Prediction.} Unlike the metrics regression framework, the scene coordinates prediction models derive the poses indirectly by predicting the key-point positions in a global coordinate system through DNNs, and calculating the camera pose using traditional projection geometry theory. \cite{massiceti2017random} \cite{li2018full} use Random Forests to infer the 3D scene coordinates corresponding to every pixel of the input image, and subsequently uses these coordinates to estimate the final camera pose via RANSAC. DSAC \cite{brachmann2017dsac} learns from SCoRF \cite{shotton2013scene} to predict scene coordinates and replaces the deterministic hypothesis selection by a probabilistic section to estimate camera pose. Any deep learning pipeline can use DSAC as a robust optimization component, like the role of RANSAC in the structure-based methods. \cite{brachmann2018learning} suggests a new, fully differentiable camera localization pipeline which has only one learnable component, a fully convolutional neural network for dense coordinate regression. SANet \cite{yang2019sanet} presents a scene agnostic neural architecture for camera localization, where model parameters and scenes are independent from each other.

\section{Proposed Approach}
\label{sec:proposed_approach}
\subsection{Overview}
\begin{figure}
	\includegraphics[width=8.5cm]{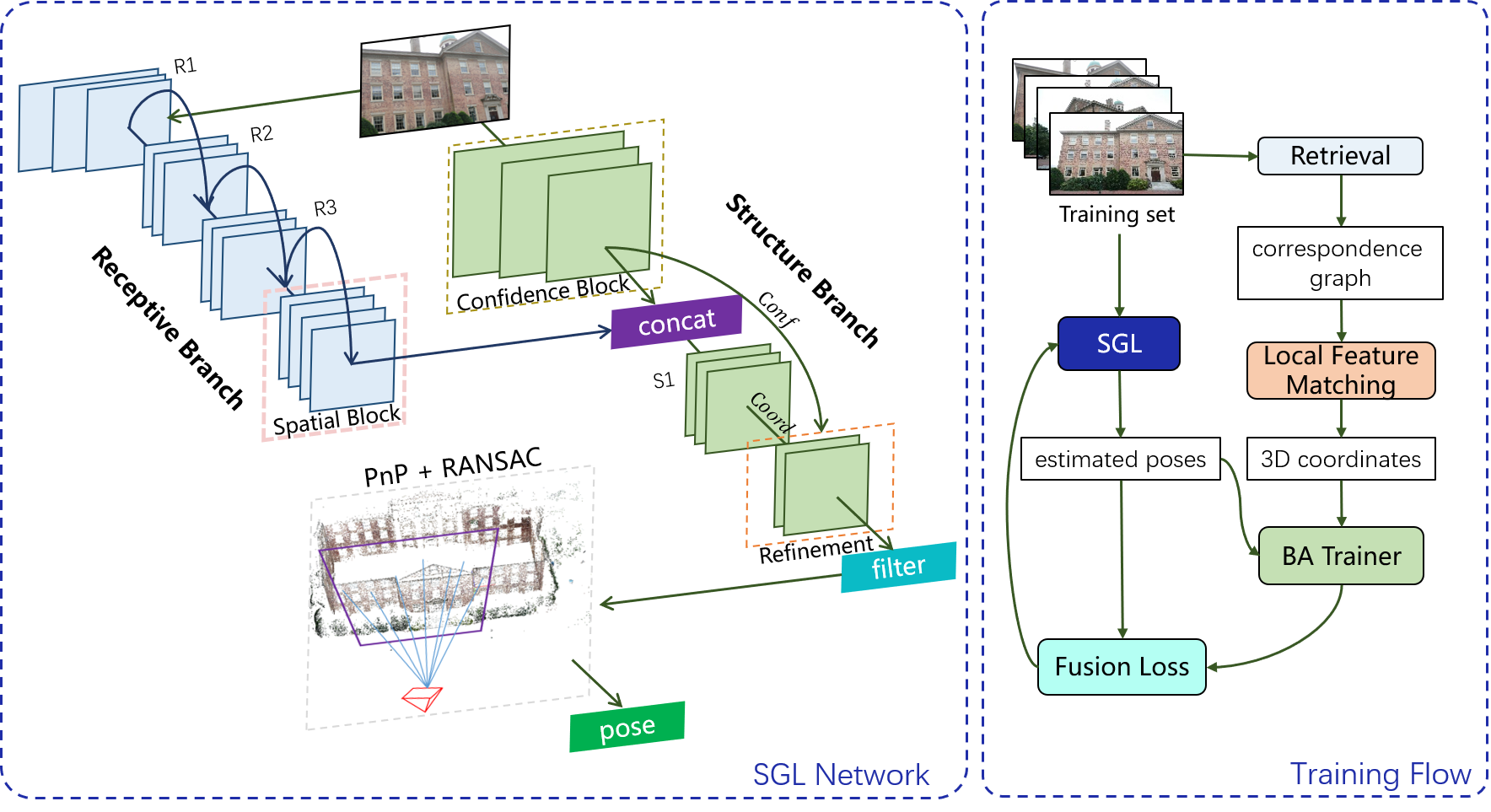}
	\centering
	\caption{
		\textbf{The Structure Guidance Learning Network structure and the training flow.} \textbf{Left:} SGL depends on the receptive branch to extract deeper context information and utilizes the structure branch to extract the shallow texture information of every input image. Then SGL fuses both feature maps and regulates them to produce fine 3D point predictions. Spatial Block is the inner component of the receptive branch, representing the complexity of the scene. Confidence Block helps to find the validity of every key-points. Finally, by applying PnP + RANSAC, SGL ensures a fine pose estimation result. \textbf{Right:} In the training stage, with the pre-trained retrieval and local feature matching model, we propose the BA trainer to fuse multi-image reprojection loss and the prediction loss for performance promotion.
	}
	\label{fig:network}
\end{figure}

In this section, we introduce the proposed method in detail. Multiple researches \cite{sattler2019understanding,huang2021vs} have revealed that structure-based approaches and the scene coordinates prediction branch of the end-to-end ones yield higher localization accuracy than the metrics regression branch of the end-to-end ones. On the other hand, although a variety of learning-based components (e.g., local feature matching, image retrieval) are introduced into structure-based methods, the main calculation burden still lies on the classical epipolar geometry computing, especially on the mapping stage. Instead, our proposed method relies on the high-performance DNNs to predict the 3D points instead of geometry computing. Namely, we substitute the SFM model with the DNN to represent the scene structure. Fig. \ref{fig:network} indicates the whole architecture of our model together with the training procedure. Table \ref{arch_table} details the layer construction of both branches.

\subsection{Receptive Branch}

Inspired by the previous solid works on the other computer vision topic \cite{yu2018bisenet,yu2021bisenet}, we also split the network into the parallel format. Receptive branch $Recep(X)$ is supposed to deal with the deeper features including the context information of the target scene and the global information with the large receptive area obtained by the cascaded multiple convolution layers. 

In this branch, we employ a Fully Convolution Network (FCN) style structure which is thoroughly testified by \cite{brachmann2018learning,brachmann2021visual} on the camera localization tasks. After the common pre-processing (e.g., normalization and dimensional reduction) of input image $X\in\mathbb{R}^{h \times w \times 3}$  with height $h$ and width $w$, the shallow layers are applied to encode the visual clues into the 1/8 feature map. The following Convolution Neural Networks (CNNs) are responsible for recovering the deeper context information within the feature map. We also create residual shortcuts \cite{he2016deep} to ensure the model convergence during the back-propagation process of the training stage. On the back-end of the Receptive Branch, there is the spatial block, which consists of four $1\times1$ kernel convolution layers $Conv$. Sufficient ablation studies in Sec. \ref{sec:ablation} implicate that its complexity is highly related to the scene content and scale. Hence, we surmise that this block is capable of embedding the spatial information of the scene into the feature map and name it with spatial block.

\subsection{Structure Branch}

Compared to the Receptive Branch, Structure Branch is responsible for extracting the low-level visual features. For that reason, the shallower network structure is more appropriate. Consequently, the 3-stacked convolution layers are adopted as the feature extractor. Meanwhile, this block is also occupied to form the confidence map $Conf_{u,v} \in \mathbb{R} ^{64\times h/8 \times w/8}$ for the further key-point refinement, therefore we mark this block as the confidence block. The channel dimension of 64 is used to remap the $8 \times 8$ down-sampled 
window-size pixels. From the confidence map, we define the pixel-wise bias $B_{i,j}\in \mathbb{R}^{2}$ by
\begin{equation} \label{B_uv}
    B_{u,v} = \mathop{\arg\max}\limits_{i,j}(Conf_{u,v}),
\end{equation}
where $i \in [0,8)$ and $j \in [0,8)$ are the index of the $8\times8$ patch, $u\in [0,w/8)$ and $v \in [0,h/8)$ are the path index within the image plane. Moreover, the confidence map also produces the absolute value of every sampled pixel, representing the reliability of the prediction. Considering the value disparity, we use the dynamic threshold (changing depending on the confidence outputs) for denying low-quality predictions in contrary to a fixed-threshold filter.
\begin{equation} \label{thresh}
    Thresh = \alpha \max_{u,v} ( Conf_{u,v}) + (1-\alpha)\min_{u,v} ( Conf_{u,v}) ,
\end{equation}    
where $\alpha \in [0,1]$ is a tunable parameter, that adjusts the domination difference between datasets. 

The output feature maps of the confidence block and the whole receptive branch are fused by the $concat$ operator. The following cascaded $Conv$ layers help to converge the channel-wise information and produce the grid-sampled 3-channel feature map $Coord \in \mathbb{R}^{3 \times h/8 \times w/8}$ representing the corresponding 3-axis global coordinate of the down-sampled pixels. However, the direct output of $Coord$ is corresponding to the sparse 2D pixel positions. Due to the resolution limits, the original output will face the wrong 2D position prediction problems. Consequently, we add an additional operator that helps us refine the 2D position on the image plane. The following equation depicts the details of the key-point position refinement and selection strategy, where $Coarse_{i,j}$ and $Fine_{i,j}$ are the original coarse 2D pixel positions and the refined positions. By using $Thresh$ we are able to filter the unreliable predictions.
{\small
\begin{equation} \label{eq:pipeline}
Fine_{u,v} = \left\{ 
    \begin{aligned} 
    Coarse_{u,v} + B_{u,v}, \quad Conf_{u,v} \geq Thresh. \\
    Null,\quad  Conf_{u,v} < Thresh.
    \end{aligned}
\right.
\end{equation}
}
With predicted 2D points and corresponding 3D predictions, we are capable of conducting PnP + RANSAC to calculate the final camera poses. 

\begin{table}[htbp]
\tiny
\centering
\caption{Instantiation of the Receptive Branch and Structure Branch. $opr$ means the operator name. $k$ is the kernel size of the convolution layer. $c$ is the output channel size. $s$ is the stride size and $p$ is the padding size. Input image sizes are different depending on the dataset, while we take $512\times640$ as an example. We use ReLU \cite{agarap2018deep} as the non-linear activation layer after most of the convolution layers. After the Refinement block, we use the BReLU-1 \cite{2017Universal} to control the value $\in [0,1)$.}
\label{arch_table}.

\renewcommand\arraystretch{1.0}
\setlength{\tabcolsep}{1.0mm}
\begin{tabular}{cccccc|cccccc|c}
\hline
& \multicolumn{5}{c|}{Receptive Branch} & & \multicolumn{5}{c|}{Structure Branch} &  Output Size  \\
                & opr& k& c &s&  p & & opr& k & c & s & p & \\
\hline
Input & -      & - & - & - &- & Input &-&-& - &- & - &  $512\times640$\\
\hline
\multirow{4}{*}{R1}     & Conv+ReLU & 3 & 32& 1 & 1&   ~  &Conv+ReLU & 3 & 32& 1 & 1& $512\times640$\\ 
                        ~& Conv+ReLU & 3 & 64& 2 & 1& Confidence &Conv+ReLU & 3 & 64& 2 & 1& $256\times320$\\ 
~& Conv+ReLU & 3 & 128& 2 & 1&  Block&Conv+ReLU & 3 & 128& 2 & 1 & $128\times160$\\ 
                        ~& Conv+ReLU & 3 & 256& 2 & 1&  ~&Conv+ReLU & 3 & 256& 2 & 1 & $64\times80$\\ 
\hline
\multirow{3}{*}{R2}     & Conv+ReLU & 3 & 256& 1 & 1&   \multirow{3}{*}{S1}  & Conv+ReLU & 1 & 512& 1 &0 & \multirow{3}{*}{$64\times80$}\\ 
                        ~& Conv+ReLU & 1 & 256& 1 & 0&  ~& Conv+ReLU & 1 & 512 &1 &0 & ~\\ 
                        ~& Conv+ReLU & 3 & 256& 1 & 1&  ~& Conv & 3 & 3 &1 &1 & ~\\ 
\hline

\multirow{4}{*}{R3}     & Conv+ReLU & 3 & 512& 1 & 1&   ~  &~  & ~ & ~ & ~ &~ & \\ 
                        ~& Conv+ReLU & 1 & 512& 1 & 0&  Refinement&  Conv & 1 & 128 &1 & 0&\multirow{2}{*}{$64\times80$}\\ 
                        ~& Conv+ReLU & 3 & 512& 1 & 1&  Block& Conv+BReLU-1 & 1 & 64& 1 & 0& ~\\ 
                        ~& Conv & 3 & 512& 1 & 1&  ~&  ~&  ~ & ~ & ~& ~& ~\\ 
\hline

~     & Conv+ReLU & 1 & 512& 1 & 0&   \multirow{4}{*}{-}  & & \multirow{4}{*}{-} & & & & \multirow{4}{*}{$64\times80$}\\ 
                            Spatial & Conv+ReLU & 1 & 512& 1 & 0&  ~&  ~&  ~ & ~ & ~& ~& ~\\ 
                            Block & Conv+ReLU & 1 & 512 & 1 & 0 &  ~&  ~& ~  &~  &~ & ~& ~\\ 
                            ~& Conv+ReLU & 1 & 512& 1 & 0&  ~& ~ &  ~ & ~ &~ & ~& ~\\ 
\hline
\end{tabular}
\end{table}

\subsection{Bundle Adjustment Training}

\begin{figure}
	\includegraphics[width=8cm]{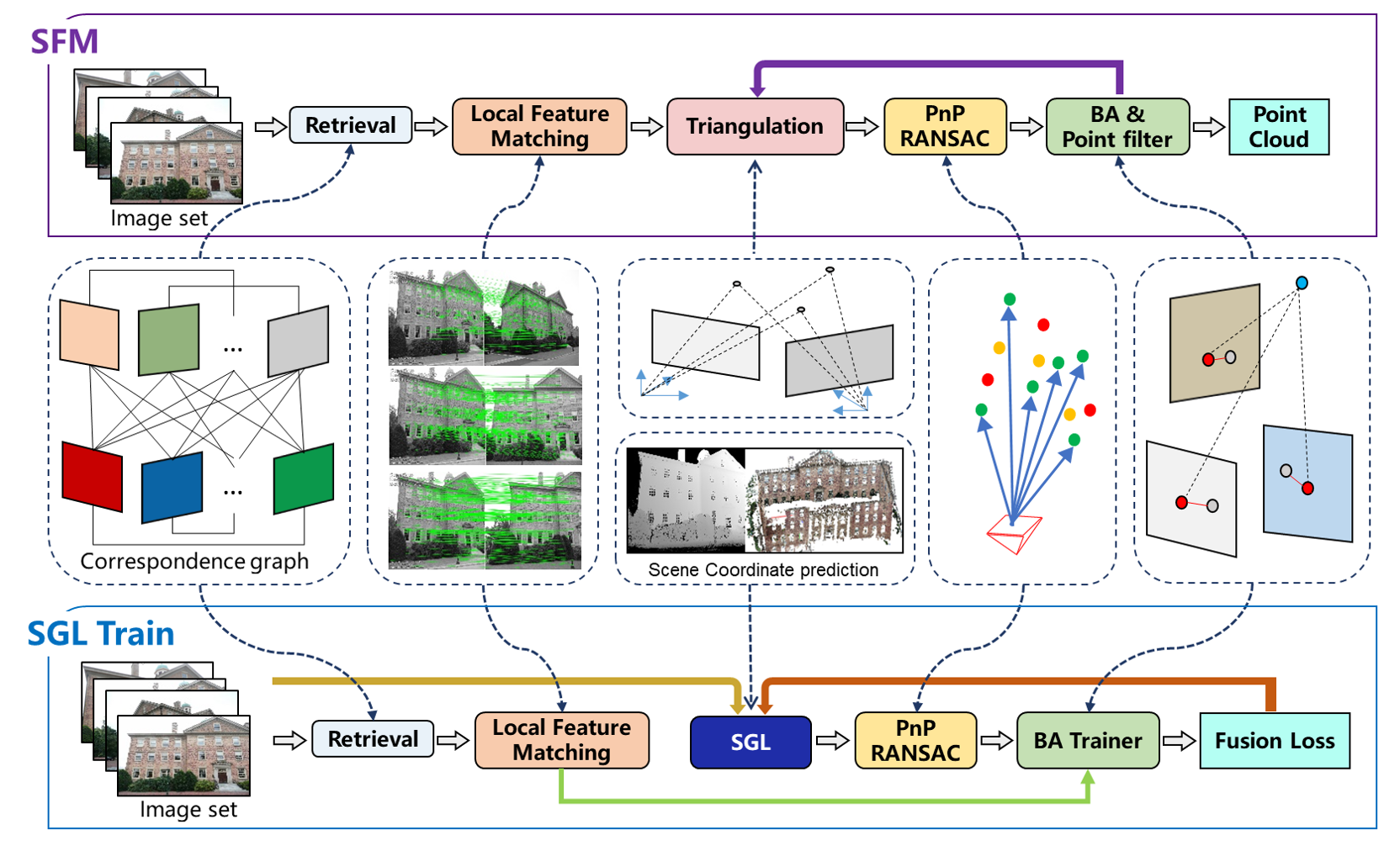}
	\centering
	\caption{
		\textbf{Comparison between a typical SFM pipeline and the proposed SGL training procedure.} \textbf{Upper represents SFM:} With the input image set, image retrieval model uses the extracted global descriptors to form the correspondence graph. The local feature extractors find the pixel-level matches of every image pair. Triangulation is based on the epipolar geometry to create the 3D points. PnP + RANSAC is used to calculate the image poses with the 2D-3D observations. Bundle Adjustment and 3D point filter help to increase the accuracy of the output point cloud circularly. \textbf{Lower represents SGL training:} Multiple components are duplicated regarding SFM, including retrieval, local feature matching and PnP + RANSAC. But SGL replaces the triangulation by the network inference. Additionally, reprojection error fused with other losses is in charge of network training instead of point cloud refinements in SFM.
	}
	\label{fig:sfm}
\end{figure}

Inspired by the traditional BA process in the SFM pipeline, we transfer BA into network training and further experiments also show its practicality. Fig. \ref{fig:sfm} thoroughly compare the algorithm components between SFM and the SGL training procedure. In the SFM pipeline, image retrieval model and local feature matching model are occupied to achieve the mapping database. With the following geometric methods, including triangulation, PnP, RANSAC and BA, we are able to produce the high accurate sparse point cloud. 

Whereas, in the SGL training pipeline, triangulation is substituted by SGL model inference. SGL achieves to get sampled 3D positions in the global frame in the meantime produce the 2D-3D pixel-level pairs. Afterwards, we take the advantage of PnP + RANSAC to calculate the camera poses. BA trainer is the training strategy that fully uses the BA idea into model training and helps the model better fit the scene. Finally, we fuse the BA loss and the metric loss and back propagate the gradients into the model weights. 

Given an image whose predicted pose is denoted by the rotation matrix $R$ and the translation vector $t$, we can calculate the pose by PnP + RANSAC solver with 2D-3D matches. Relatively, the ground truth poses are rotation matrix $\overline{R}$ and translation vector $\overline{t}$. The metric loss including angle loss $\mathcal{L}_{angle}$ and translation loss $\mathcal{L}_{trans}$ can be obtained by following two equations.
\begin{equation} \label{loss_rotation}
    \mathcal{L}_{angle} = \frac{1}{\pi}\arccos{(\frac{1}{2}trace(\overline{R}^T \cdot R - 1))}.
\end{equation}
\begin{equation} \label{loss_fuse}
    \mathcal{L}_{trans} = \Vert \overline{t} - t \Vert_2.
\end{equation}
Subsequently, we define the BA loss $\mathcal{L}_{BA}$ by the following equation.
{\footnotesize
\begin{equation} \label{loss_repro}
    \mathcal{L}_{BA} = \sum\limits_{i=0}^{n}\left[{\Vert p_i - K \cdot \left(R_i \cdot y_i + t_i\right) \Vert_2 + ( \frac{1}{n}\sum\limits_{j=0}^{n}{y}_{j}-y_i)}\right],
\end{equation}
}
where $i$ is the index number in the total $n$ bundle images. $p_i$ represents the 2D positions on the image plane of the sampled pixels, while $y_i$ is the predicted corresponding 3D position vector in the global frame. $K$ is the $3 \times 3$ camera intrinsic matrix. By applying the SGL predicted rotation $R_i$ and translation $t_i$, we are able to calculate the reprojection errors of every image in the bundle set. Considering the instability of the SGL inference, we use the mean coordinates to compute the variance into the loss, in order to stabilize the network outputs. Consequently, we synthesize all the loss elements together, illustrated by the following equation.
\begin{equation} \label{loss_repro}
    \mathcal{L}_{all} = \mathcal{L}_{angle} + \beta \mathcal{L}_{trans} + \gamma \mathcal{L}_{BA},
\end{equation}
where $\beta $ and $\gamma$ weigh the contributions of the different loss components.


\section{Experimental Evaluations}
\label{sec:experimental_evaluations}

\subsection{Datasets}

We quantify the visual localization performance of SGL on two public benchmark datasets Microsoft
7-Scenes \cite{shotton2013scene} (indoor) and Cambridge Landmarks Dataset \cite{kendall2015posenet} (outdoor). Both datasets contain challenging conditions, including motion blur, large illumination change, dynamic obstacles and etc. Ground truth poses are provided by the higher accurate equipments, which are the Kinect RGB-D camera in 7-Scenes and the abundant SFM data in Cambridge Landmarks. Besides, both benchmark datasets are widely occupied for visual localization evaluation. Many state-of-the-art localization researches have reported their statistic results, which makes them the validate baselines.

\subsection{Implementation Details}
Hyperparameters $\beta=100$ and $\gamma=0.25$ are set fixed, while $\alpha$ varies from $0.70$ to $0.95$ on different subsets, which will be discussed in Sec. \ref{sec:ablation}. The following experiments are conducted on the NVIDIA Tesla V100 GPU with the CUDA 10.2 and Intel(R) Xeon(R) Gold 6142 CPU @ 2.60GHz. 

\subsection{Comparison with State-of-the-art Approaches}

\begin{figure}
	\includegraphics[width=9cm]{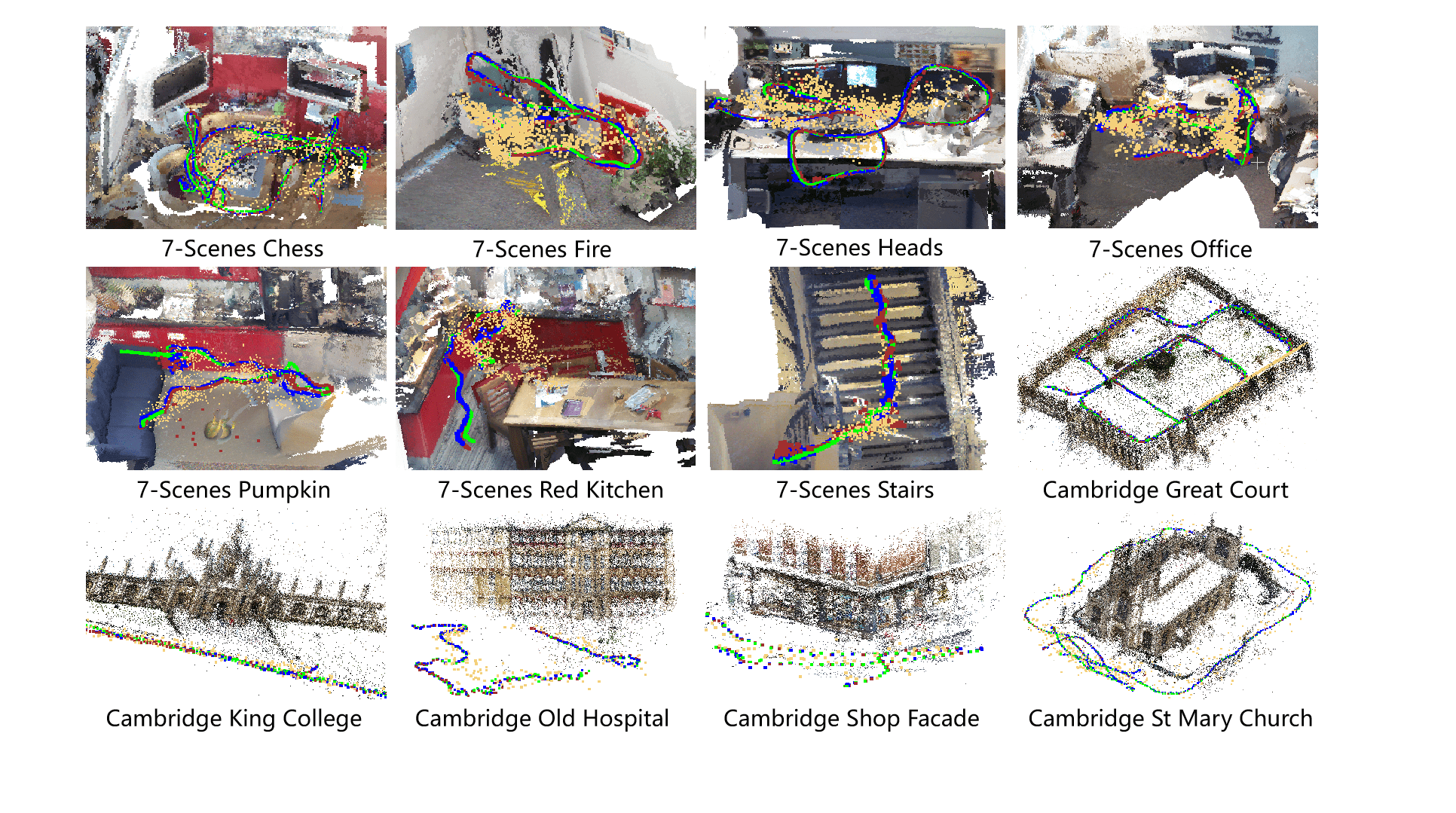}
	\centering
	\caption{
		\textbf{Localization Trajectories of 7-Scenes and Cambridge.} The green trajectory is the ground truth. The blue one is the SGL predictions. The red one is HF-net predictions. The yellow one is the PoseNet predictions. On most of the subsets, SGL's predictions are closer to the ground truth than the other two methods.
	}
	\label{fig:traj}
\end{figure}

In quantitative analysis, we select two of the representative visual localization methods among each branch discussed before. Statistics show that structure-based and coordinates prediction methods perform better than metrics regression ones. Because the coordinates prediction methods benefit from the high performance of the DNN and achieve the better performance than structure-based ones. Furthermore, our proposed SGL network outperforms most methods on both indoor and outdoor benchmark datasets, with higher translation and rotation accuracy. In Fig. \ref{fig:traj} we also demonstrate the localization trajectory of those datasets to prove the accuracy. 

\begin{table}[htbp]
\renewcommand\arraystretch{1.2}
    \scriptsize
	\centering
	\caption{Localization accuracy comparison with SOTA approaches. The median translation error (cm) and angular error (\degree) of each subset are reported, but the dash (-) sign represents the failure of the certain method on the certain subset.}
	\label{sota}
    \setlength{\tabcolsep}{1.8mm}{
	\begin{tabular}{ccccccc}
	\hline
	    \multirow{2}{*}{ } & \multicolumn{1}{c}{S.} & \multicolumn{1}{c}{M. R.} & \multicolumn{3}{c}{C. P.} \\
    \cline{2-6}
	 ~&  HLoc \cite{brachmann2021limits} & MS-T. \cite{shavit2021learning}& DSM \cite{tang2021learning} & DSAC* \cite{brachmann2021visual} & SGL \\
	\hline
	Chess & 2.4, 0.77 & 11, 4.66 & 2, 0.71 & 1.9, 1.11 & 1.5, 0.52 \\
    Fire & 1.8, 0.75 & 24, 9.6 & 2, 0.85 & 1.9, 1.24 & 1.8, 0.67 \\
    Heads & 0.9, 0.59 & 14, 12.19 & 1, 0.85 & 1.1, 1.82 & 1.2, 0.78 \\
    Office & 2.6, 0.77 & 17, 5.66 & 3, 0.84 & 2.6, 1.18 & 2.3, 0.68 \\
    Pumpkin & 4.4, 1.15 & 18, 4.44 & 4, 1.16 & 4.2, 1.41 & 3.2, 0.94 \\
    Kitchen & 4.0, 1.38 & 17, 5.94 & 4, 1.17 & 3.0, 1.70 &  2.9, 0.94 \\
    Stairs & 5.1, 1.46 & 26, 8.45 & 5, 1.33 & 4.1, 1.42 & 3.1, 1.21 \\
    \hline
	\end{tabular}
	}
\end{table}

\subsection{Ablation Studies and Discussions}
\label{sec:ablation}
\begin{figure}
	\includegraphics[width=7cm]{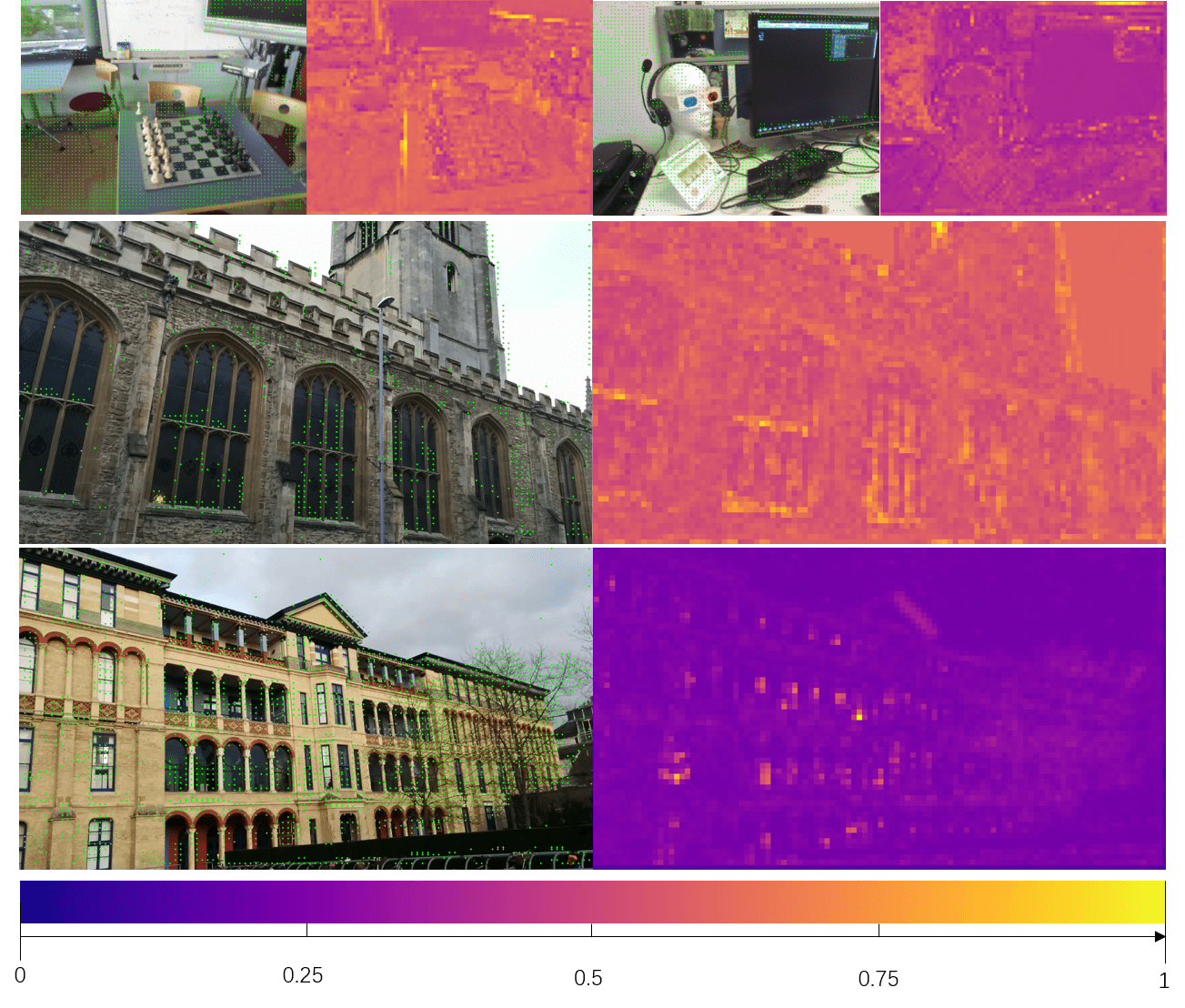}
	\centering
	\caption{
		\textbf{The Saliency Map of confidence map $Conf$.} We provide Chess, Heads, Church and Hospital subsets in 7-Scenes and Cambridge datasets, in order to prove the performance of the confidence map and the filter strategy. The green points are the surviving key-points with high confidence after filtering. 
	}
	\label{fig:saliency_map}
\end{figure}

\textbf{Confidence Block.} To illustrate the validity of the confidence block, we visualize the output of the confidence block in Fig. \ref{fig:saliency_map}. The high values of the down-sampled map represent the high confidence, and vice versa. Eq. \ref{thresh} ensures us to select the high-confidence points with high-quality observation. Qualitatively, with the help of key-point filter and refinement, the green points focus mostly on the objects with meaningful textures, such as architectures, cabinets and posters. Points on the objects with less texture information, like the sky, black screen and windows are neglected by our strategy. Moreover, although green points are formatted in a grid-like way because of the down-sample effect by the CNNs, some points that lie near the object edges are tunned to fit the contours, because of Eq. \ref{B_uv}.

\begin{figure}
	\includegraphics[width=8.5cm]{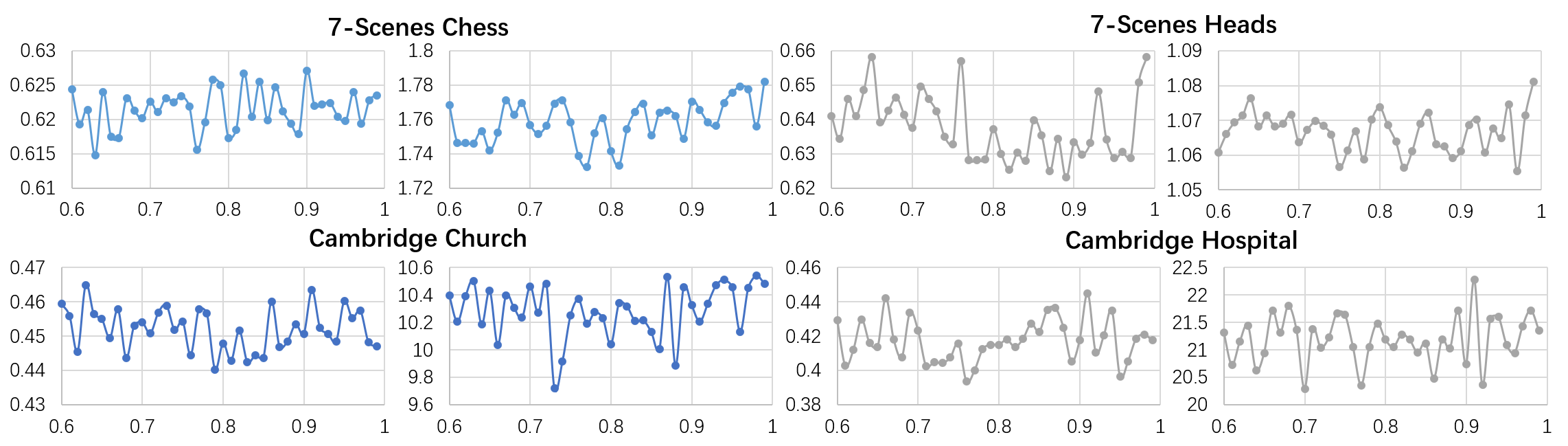}
	\centering
	\caption{
		\textbf{Alation study on confidence ratio $\alpha$.} We provide the error distribution curves on Chess, Heads, Church and Hospital subsets in 7-Scenes and Cambridge datasets. The $x$-axis is the value of $\alpha$. The $y$-axis is the median rotation error (\degree) on the left and the median translation error (cm) on the right of every scene. 
	}
	\label{fig:alpha}
\end{figure}

Correspondingly, due to the instability and inexplicability of the DNNs, the outputs of the confidence block vary greatly under different circumstances on a single sequence. Therefore instead of a fixed threshold, a dynamic threshold would avoid uneven remaining points circumstances caused by the unreasonable threshold. Fig. \ref{fig:alpha} provides the ablation study of the localization accuracy with respect to the tunable confidence ratio $\alpha$. As $\alpha$ balances the number of SGL predictions, therefore medium number for preserved key-points shows better performance on localization precision. Correspondingly, Fig. \ref{fig:alpha} shows $\alpha$ that lies within 0.70~0.95 holds lower rotation and translation errors. Because too few observations will result in noise in the PnP process, and too large $\alpha$ leads to failures in excluding those low-quality predictions of SGL.

\begin{figure}
	\includegraphics[width=6.5cm]{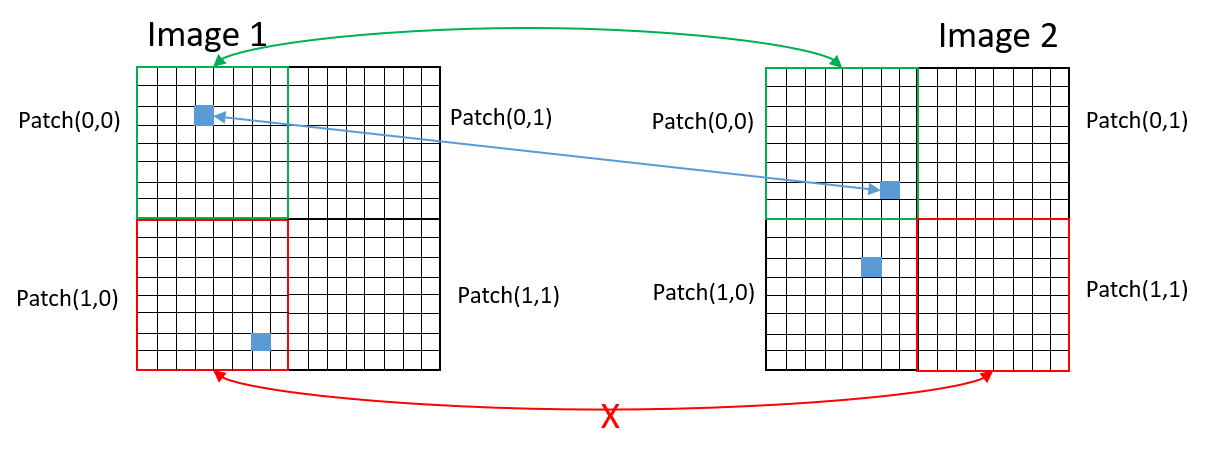}
	\centering
	\caption{
		\textbf{The pixel-wise adjustment strategy in the BA trainer.} For simplicity, we only show $2\times2$ patches per image. Every patch contains $8\times8$ pixels. The blue pixel is the SGL tunned key-point position in the relative patch. The red curve means the eliminated SuperGlue match considering the filtered position by SGL. The Green curve means the remained patch-wise match pair. The blue line means the remained pixel-wise match pair which is fused with SGL and SuperGlue outputs.
	}
	\label{fig:ba}
\end{figure}

\textbf{Bundle Adjustment Trainer.} We occupy HOW \cite{tolias2020learning} to achieve the images corresponding graph and SuperGlue \cite{sarlin2020superglue} to acquire the local feature matches. Nevertheless, the original output of the local feature networks do not always share the same key-points with the SGL. As a result, we have to adjust the matching pairs to fit the SGL key-point predictions. In the BA training process, we fully trust the matches produced by SuperGlue. As shown in Fig \ref{fig:ba}, we downsample the original local feature matches produced by SuperGlue by $8\times8$ pixels per patch. Namely, every patch only preserves a single match. On the other hand, SGL provides the key-point pixel-wise position in every patch. Fig. \ref{fig:ba} shows the circumstances when the key-point is tunned, modified and even filtered during the training process. 

\begin{figure}
	\includegraphics[width=8.5cm]{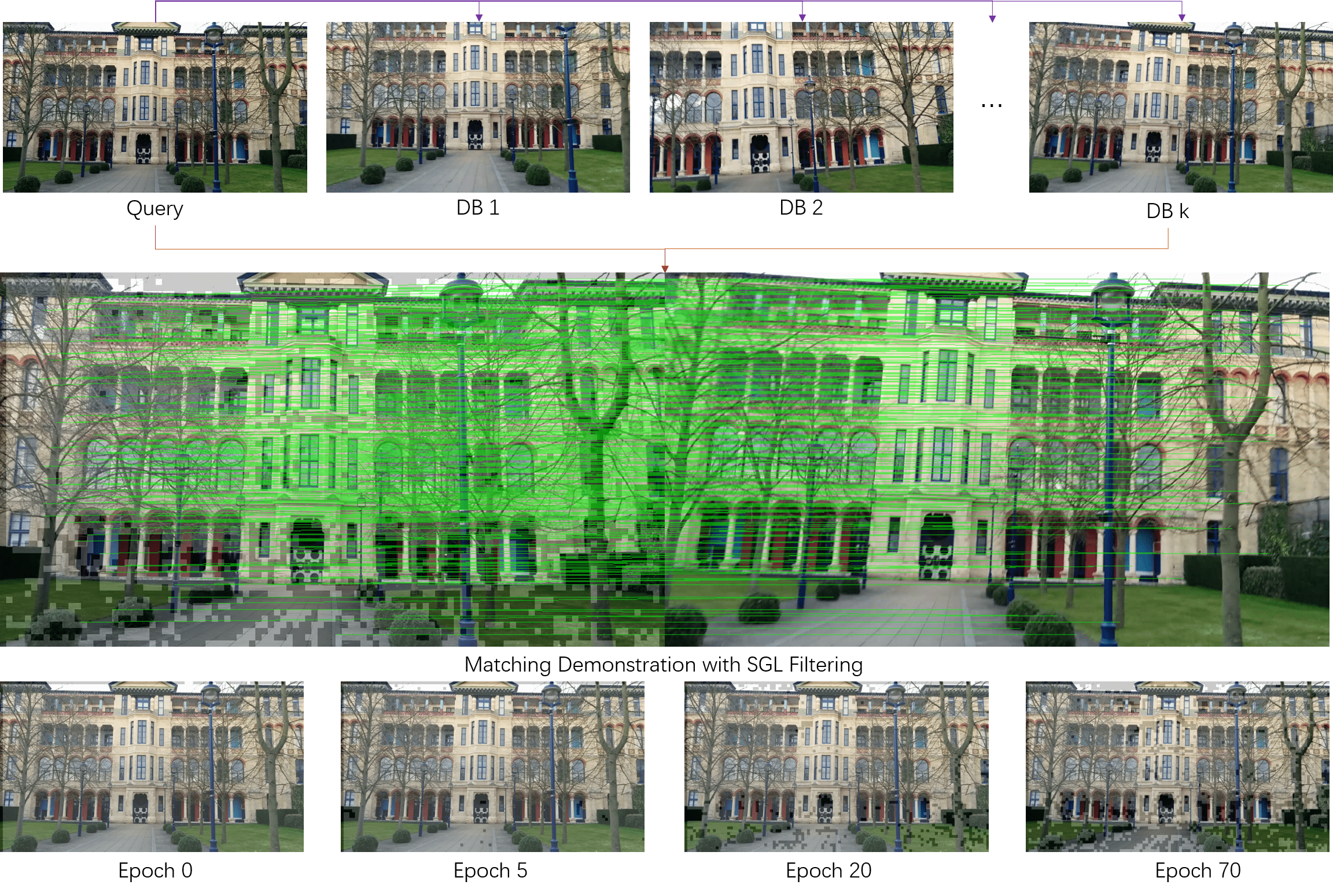}
	\centering
	\caption{
		\textbf{Visualization of the influences from the confidence filter and BA trainer.} The top row shows the training datasets with one query image and k retrieved database images produced by HOW. The mid row shows the matching results (Query V.S. DB k) filtered by the SGL confidence module. The dark mosaics are the patches whose confidence scores are lower than the threshold. The bottom row shows the confidence filtering progress during the training stage. 
	}
	\label{fig:ba_matching}
\end{figure}

Fig. \ref{fig:ba_matching} illustrates the training status of the SGL network. With retrieved $k$ candidates and local feature matches, we are able to establish the key-point level correspondences in every bundle of training images. SGL's confidence module predicts the high-confidence patches and eliminates the low-confidence ones, which are represented by the dark mosaics in the mid and bottom rows of the Fig. \ref{fig:ba_matching}. As the training carries forward, the confidence module will gradually fit the scene and locate the low texture patches to exclude them from the later training stage. Table. \ref{table:ablation} interprets that with the BA trainer's help the localization accuracy rises on both datasets. 

\textbf{Spatial Block.} In the Sec. \ref{sec:proposed_approach} we introduce the spatial block in the receptive branch, because the depth of this block is sensitive to the final localization performance in different scenes, according to the statistics in Table. \ref{table:ablation}. By increasing the $Conv$ layer number of the spatial block, we can improve the precision of SGL on the most subsets of 7-Scenes (except Stairs), while on the Cambridge dataset such phenomenon goes oppositely. Typically, when we substitute the spatial block with a much more complicated block, Transformer \cite{vaswani2017attention}, the accuracy polarized greatly on these scenes. But the trend of accuracy transfers synchronizes with the ablation study on the changes of Conv layer number. In some subsets of both datasets, the higher complexity of spatial block has great effects on precision, while its performance is not stable especially on Cambridge datasets. Through all the ablation experiments we attribute the sensitivity of the spatial block to the diversity of the scenes on different subsets. 

\begin{table}[htbp]
\renewcommand\arraystretch{1.2}
    \tiny
	\centering
	\caption{Ablation study on BA Trainer and Spatial Block. S. represents the number of layers in the spatial block. The transformer block represents the network structure described in \cite{vaswani2017attention}. The median translation error (cm) and angular error (\degree) are reported.}
	\label{table:ablation}
	\begin{tabular}{cccccc}
    \hline
	 ~ & SGL (S.=4) & SGL (S.=4, w/o BA) & S.=2 & S.=6 & S.=Transformer \\
	\hline
	Chess & 1.6, 0.77 & 1.8, 0.83 & 1.7, 0.71 &  \textbf{1.6}, 0.59 &  \textbf{1.6}, \textbf{0.52}\\
    Fire & 1.9, 0.94 & 2.0, 0.95& 1.9, 0.94 & 1.9, 0.79  &  \textbf{1.8}, \textbf{0.67} \\
    Heads & \textbf{1.1}, 0.84 & 1.2, 0.88& 1.3, \textbf{0.72} & 1.2, 0.75 & 1.3, 0.79\\
    Office & 2.6, 0.74 &2.8, 0.82 & 2.7, 0.78 & 2.5, 0.71 & \textbf{2.4}, \textbf{0.68}\\
    Pumpkin & 3.9, 1.04 &3.9, 1.10 & 4.0, 1.09 & 3.7, 1.03 & \textbf{3.2}, \textbf{0.94} \\
    Kitchen & 3.8, 1.20 &4.0, 1.24 & 3.9, 1.29 & 3.7, 1.12& \textbf{3.0}, \textbf{0.95}\\
    Stairs & \textbf{3.9}, 1.06 & 4.4, 1.22 & 4.1, 1.08 & 5.1, 1.54 & 11.2, 1.77\\
    \hline
    Great Court & \textbf{27.5}, \textbf{0.21} & 32.1, 0.25  & 30.6, 0.31 & 35.1, 0.39 & 59.0, 0.41 \\
    Kings College & \textbf{12.9}, \textbf{0.32} & 16.7, \textbf{0.32} & 16.2, 0.42 & 15.8, 0.41 & 20.4, \textbf{0.32}\\
    Old Hospital & 22.9, \textbf{0.41} & 21.5, 0.60 & \textbf{20.2}, 0.55 & 26.9, 0.60 &30.7, 0.57\\
    Shop Facade & \textbf{5.1}, 0.28 & 5.6, \textbf{0.27} & 5.2, 0.28 & 6.2, 0.31  &37.5, 1.34\\
    St M. Church & \textbf{10.5}, \textbf{0.44} & 12.1, 0.51 & 12.5, 0.59 & 14.8, 0.67 & 21.0, 0.75\\
    \hline
	\end{tabular}
\end{table}

\section{Conclusions}

In this paper, we present an end-to-end framework camera localization network, SGL, which takes the advantage of DNN to predict the scene coordinates of the sampled key-points and utilizes PnP+RANSAC to estimate the predicted camera poses. To accomplish better accuracy, we divide the network into two branches, receptive branch and structure branch, in order to fuse the low-level feature and high-level one. Moreover, we design the key-point refinement strategy to escalate the precision of 3D key-point observations. In the training part, we introduce BA trainer to fully use the multi-view information to help the network fit the scene. We conduct sufficient experiments and compare our method to some SOTA ones. Statistics prove the validity and high precision of the proposed method. Our further works will concentrate on adapting SGL to the sequential inputs and exploring the probability to simplify the BA trainer by omitting the extra relyments on the image retrieval and the local feature matching.









\bibliographystyle{ieeetr}
\bibliography{sgl}

\end{document}